\documentclass[sigconf]{acmart}

\usepackage{booktabs} 
\usepackage{multicol}

\setcopyright{rightsretained}

\acmDOI{10.475/123_4}

\acmISBN{123-4567-24-567/08/06}

\acmConference[WOODSTOCK'97]{ACM Woodstock conference}{July 1997}{El
  Paso, Texas USA}
\acmYear{1997}
\copyrightyear{2016}

\acmArticle{4}
\acmPrice{15.00}

\editor{Jennifer B. Sartor}
\editor{Theo D'Hondt}
\editor{Wolfgang De Meuter}

\begin{document}
\title{Prediction and Localization of Student\\ Engagement in the Wild }

  \author{Amanjot kaur, Aamir Mustafa, Love Mehta, Abhinav Dhall}

\begin{abstract}
In this paper, we introduce a new dataset for student engagement detection and localization. Digital revolution has transformed the traditional teaching procedure and a result analysis of the student engagement in an e-learning environment would facilitate effective task accomplishment and learning. Well known social cues of engagement/disengagement can be inferred from facial expressions, body movements and gaze pattern. In this paper, student's response to various stimuli videos are recorded and important cues are extracted to estimate variations in engagement level. In this paper, we study the association of a subject's behavioral cues with his/her engagement level, as annotated by labelers. We then localize engaging/non-engaging parts in the stimuli videos using a deep multiple instance learning based framework, which can give useful insight into designing Massive Open Online Courses (MOOCs) video material. Recognizing the lack of any publicly available dataset in the domain of user engagement, a new `in the wild' dataset is created to study the subject engagement problem. The dataset contains 195 videos captured from 78 subjects which is about 16.5 hours of recording. We present detailed baseline results using different classifiers ranging from traditional machine learning to deep learning based approaches. The subject independent analysis is performed so that it can be generalized to new users.  The problem of engagement prediction is modeled as a weakly supervised learning problem. The dataset is manually annotated by different labelers for four levels of engagement independently and the correlation studies between annotated and predicted labels of videos by different classifiers is reported. This dataset creation is an effort to facilitate research in various e-learning environments such as intelligent tutoring systems, MOOCs, and others.  
\end{abstract}

%
%



\maketitle
\section{Introduction}
\label{sec:intro}
Human beings require time and persistent effort to perform a task such as learning. On the same lines, the researchers have argued the importance of \emph{continuous effort or engagement to accomplish the learning task} \cite{nicholls2015some}. In any education system, student engagement is a key component.  With the advent of digital technologies, the traditional classroom teaching is transformed into advanced learning environments such as an Intelligent Tutoring System (ITS) and e-learning environment such as Massive Open Online Courses (MOOCs). As a result, Student engagement in the learning environment has also adapted in its own way.\\
\emph{Student engagement} is defined as a complex structure with multi-dimensions and components. 
Various other works have classified it in different ways.
\begin{figure}[t]
\centering
 \includegraphics[width=0.48\textwidth]{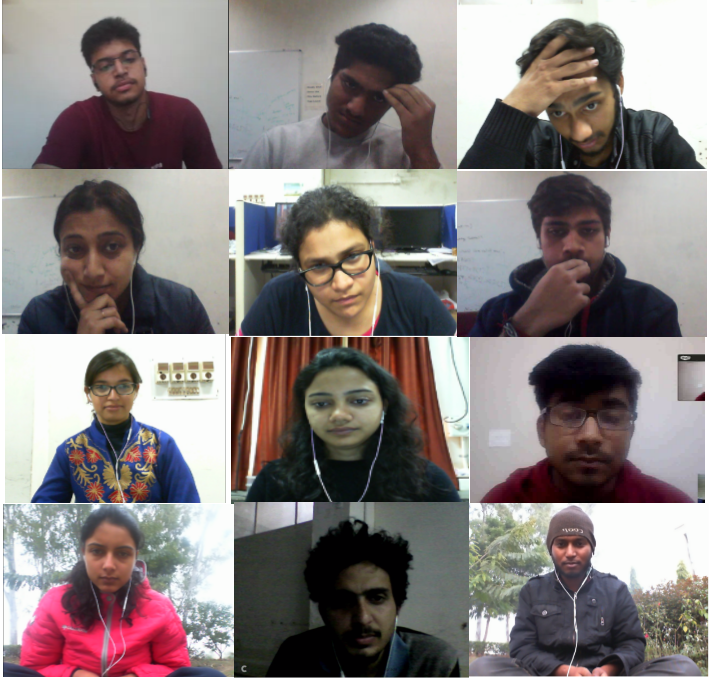}
\caption{\small Examples of frames from our engagement database. Top to bottom rows show engagement intensity level: [0 (low) - 3 (high)].   } \vspace{-2mm}
\label{fig:dataset}
\end{figure}
Fredricks \emph{et al.} proposed three components \cite{fredricks2004school} as \emph{Behavioral Engagement} component which explains students involvement in terms of effort, persistence and concentration. \emph{Emotional Engagement} is related to feelings of interest or attitude towards a particular theme.\emph{ Cognitive Engagement} focus on allocation of effort, a strategy used, in terms of cognitive effort, for the accomplishment of the task. Other models have introduced another dimension known as \emph {Agentic Engagement} emphasize on proactive actions taken by the student for learning a particular task \cite{reeve2011agency}.\\
Several traditional methods and measures introduced in literature to assess the level of engagement have their own relevance. Different methods such as self-reports: participants answer the set of questions related to their experience such as level of engagement, interest and so on \cite{grafsgaard2012multimodal}.
Secondly, an observational study was done by the external expert: Focus on the behavioral analysis of students.\\
Traditional measurement methods are not sufficient to measure engagement in all the contexts. So automatic engagement assessments for the digitally transformed learning environment are required. These techniques analyze various facial cues, body posture, well known social cues of engagement and disengagement captured automatically using affective computing techniques. These techniques are sensitive to the engagement levels variations over time. Moreover, automatic measures can facilitate timely intervention to change the course of fading engagement level \cite{d2017advanced}. 
\\
It is evident that student engagement plays a key role in the learning process, so its automatic measurement is also important in today's digital learning environment.
\emph{The paper proposes an automatic pipeline for predicting user's engagement, while they are watching educational videos such as the ones in Massive Open Online Courses (MOOCs)}. \emph{Secondly, A new ` Engagement Detection in the wild' video-based database (Figure \ref{fig:dataset}) is also introduced}. User engagement analysis is of importance to learning environments, which involves interaction between humans and machines. Generally, a course instructor is able to assess the engagement of students in a classroom environment based on their facial expression. Other well known social cues such as yawning, glued eyes, and body posture are also handy in analyzing the behavioral change of the participants in the natural environment.  However, this assessment becomes tricky, when students are watching study material such as videos, on their computers and mobile phones in different environments. These different environments can be their room, office, cafeteria, outdoor and so on.    


\emph{Engagement} is defined as a connection between a user and a resource, which comprises of emotional, cognitive and behavioral nature at any point in time \cite{wu2017beyond}. This work is related to the estimation of perceived engagement by the external observer and it is appropriate for the e-learning environment. Various examples of environments in which the engagement of user is a vital component to analyze are: e-learning, health sector such as conducting autism-related studies, analyzing driver engagement in automated cars and so on.\\
Automatic engagement prediction can be based on various kinds of data modalities. It is argued that the student response can be used as an indicator of engagement in intelligent tutoring systems \cite{koedinger1997intelligent}. Another approach is based on features extracted on the basis of facial movements \cite{d2010multimodal,xiao2017dynamics}. Automated measures such as response time of a student to problems and test quizzes are used in ITS \cite{joseph2005engagement}. Physiological and neurological measures, requiring specialized sensors, such as electroencephalogram, heart rate, and skin response have been used to measure engagement \cite{goldberg2011predicting, chaouachi2010affect, xiao2017undertanding}.

\begin{table}
  \caption{Interesting works for engagement detection.}
  \label{table:Literature}
  \begin{tabular}{|l| c| c| c|}
  \hline
    {\small\textbf{Name}}
    & {\small \textbf{Features}}
      & {\small \textbf{Subj.}}
    & {\small \textbf{Method}} \\
    \hline
   D'Mello et. al\cite{D'Mello} &FACS \& Log  &28   & \\ \hline
   Grafsgaard et. al\cite{grafsgaard2012multimodal} &FACS  & 67 &Regression   \\\hline
    Whitehill et al\cite{whitehill2014faces} &FACS  &34 &Regression \\\hline
   Grafsgaard et. al\cite{grafsgaard2011predicting} &FACS, Log &-&HMM \\\hline
   Gupta et. al\cite{gupta2016daisee} &Facial Features &112&DNN \\ \hline
   Xian et. al\cite{xiao2017undertanding} &PPG &18&RBF+SVM \\ \hline
   Arroyo et. al\cite{beal2007line} &Video &60&Regression \\ 
& \&Physiological & & \\ \hline
Bosch et. al\cite{bosch2016detecting} &Facial Features&137&Regression \\
&Body Posture& &\\ \hline
     
  \end{tabular}
\end{table}





In the literature (Table \ref{table:Literature}), work related to engagement detection is presented. Xiang \emph{et al.} have studied the impact of distractions or multitasking in MOOCs taken on Mobile apps \cite{xiao2017undertanding}. In this work, a total of 18 subjects watched two stimuli as lecture videos. Post the lecture, self-reports and quiz related to video content was asked. Along with this Photoplethysmography signals were recorded.\\ In an interesting work, Whitehill \emph{et al.} analyzed the behavioral engagement and defines four level of engagement. The analysis is performed on the Facial Features (FACS) with the help of SVM Classifier \cite{whitehill2014faces}. Affective states pertaining to engagement such as Interest, Confidence, Excitement and Frustration (1-5 scale) were used. In this experiment, video and physiological sensor signals were used to capture different modalities to analyze affective states of students while solving mathematical problems of SAT. Dataset was self-annotated and linear regression was used to regress the value of different affective states \cite{beal2007line}. In another study conducted by Bosch \emph{et al.} in the real world settings, various affective states of engagement such as boredom, engaged, concentration, confusion, frustration and delight were used. Features were in the form of facial expressions and body movements. Regressors such as BayesNet, clustering, regression were used to regress different dimensions of the network \cite{bosch2016detecting}.\\
Surveys such as \cite{Zeng_PAMI_2009_short} \cite{corneanu2016survey} discuss the important work done in the field of affective computing. Based on the knowledge in affective computing, researchers have recently tried to predict user engagement. El Kaliouby \emph{et al.} \ \cite{el2006affective} argue that user engagement prediction is a challenging problem and has many applications in autism-related studies. Furthermore, Tawari \emph{et al.}\ \cite{tawari2014looking} found that a car driver's affective state recognition in terms of the engagement level provides important cues in detecting drowsiness levels while driving. Navarathna \emph{et al.}\ \cite{navarathna2017estimating} proposed an interesting pipeline for analyzing a movie's affect by predicting viewer's engagement in the movie. User engagement level recognition is inevitable in e-learning and is useful in detecting states such as fatigue, lack of interest and difficulty in understanding the content.


Sinatra \emph{et al.} has discussed person oriented perspective engagement. It revolves around states such as the cognitive, affective and motivational at the time of learning\cite{sinatra2015challenges}. These states are captured using physiological signals, facial expressions and so on\cite{d2017advanced}. Engagement has temporal dimension and dynamic in nature. Temporally constrained nature signifies the fact that it doesn't remain same, moreover, varies between low and high. 
The vital aspect here is to study the pattern of user engagement across online video lectures. It can give various interesting insights such as: for how long users concentrate, at which point in a lecture video the user looses interest etc. This is useful for successful delivery of educational content and deciding evaluation criteria for user engagement\cite{wu2017beyond}. In the research community many datasets are released to study the problems such as object detection\cite{deng2009imagenet} and segmentation\cite{martin2001database} in an image and video. The number of datasets related to user engagement in the online learning environment released is very limited \cite{attfield2011towards,sun2016bnu,whitehill2014faces}. With the advent of deep learning frameworks, larger sized databases representing diverse settings are required.

The availability of datasets related to user engagement especially student engagement in online courses would help in understanding the problems faced by students such as: loss of interest, fatigue, boredom etc in the online learning environment. A student engagement database is proposed in the report \cite{gupta2016daisee}.

In this paper, we introduce a database for common benchmarking and development of engagement assessment in diverse and `in the wild' conditions which here refer to different background, illumination and poses etc. Video level engagement prediction task is modeled as a weakly labeled learning problem. Multiple Instance Learning (MIL) is followed as the paradigm for this weakly supervised learning. MIL has been found extensively useful for tasks, in which the dataset has weak or noisy labels \cite{amores2013multiple} and has been recently used in affect analysis tasks such as facial expression prediction \cite{tax2010detection} and image emotion prediction \cite{rao2016multi}. MIL assumes that the data could be presented with the help of a set of sub-instances or bag of sub-instances as $\{X_i\}$, where each bag consists of sub-instances $\{x_{ij}\}$ and the weak labels $\{y_{i}\}$ are present at the bag level only \cite{wu2015deep}. 

Our work is based on Deep Multi-Instance Learning (DMIL) framework which makes use of complex features that can be learned by deep neural networks along with MIL technique for a weakly supervised problem. This technique has been successfully used in vision tasks like image/video classification, object detection \cite{wu2015deep,karpathy2014large}, medical image processing in the form of mammography for breast cancer detection \cite{zhu2017deep} and molecular activities of drugs \cite{dietterich1997solving}. Sikka \emph{et al.} \cite{sikka2013weakly} studied pain localization in a video in a weakly supervised fashion using the MIL framework. Tax \emph{et al.} \cite{tax2010detection} proposed an MIL based framework to select the concept frames using clustering multi-instance learning.

The main contribution of this work is to formulate student engagement prediction and localization as an MIL problem and derive the baseline scores based on DMIL. The rationale behind MIL is that labeling of engagement at frequent intervals in user videos is expensive and noisy. 

The rest of the paper is organized as follows: In Section \ref{sec:data} we introduce the dataset and details about data collection and annotation. In Section we discuss the methodology used for student engagement prediction and localization. In Section  and we describe the experimental setup and the results obtained respectively. In Section \ref{sec:experiments} and \ref{sec:results} we describe the experimental setup and the results obtained respectively.

\section{Data Collection}
\label{sec:data}

In this section, we discuss the data recording paradigm used in this work. The experiment has two parts: i) Video recordings of the subjects, while they watch the stimuli videos (each video around 5 minutes long) and ii) A verbal feedback section, in which the subject is given 15 seconds to speak about the video. In those 15 seconds the subjects are asked to express their views about the video, especially if they found the video interesting?, would they like to watch the video again?, which part of the video was most/least engaging?, any comments/suggestions on how the video could have been made more engaging etc. Based on the content of the stimuli videos, we choose  purely educational videos \emph{Learn the Korean Language in 5 minutes}, \emph{a pictorial video (Tips to learn faster)} and  \emph{(How to write a research paper)}. This was aimed to capture both focused and enjoyable settings which allow natural variations. In the experiment section, as a baseline, we explore the video signals only.

The dataset has 78 subjects (25 female and 53 male) in total. The age range of the subjects is 19-27 years. A total of 195 videos are collected, each approximately 5 minutes long. The dataset is collected in unconstrained environment i.e. at different locations such as computer lab, hostel rooms, open ground etc. In order to introduce the effect of different environments, we also include a video conferencing based setup. In this, the subject, watched the stimuli on their computer (full) screen and in parallel a Skype-based video session was setup. We captured the video region of the Skype application at the other end of the Skype video-call. It was made sure that the subject was not disturbed by the Skype recording. The usefulness of recording over Skype is that network latency and frame drop can help in simulating different environments. 

The videos are captured at a resolution of 640 x 480 pixels at 30 fps using a Microsoft Lifecam wide-angle F2.0 camera (excluding the Skype-based recordings). The audio signal is recorded using a wideband microphone, which is part of the camera. Apowersoft Screen Recorder software is used to capture the Skype videos. 

The dataset and it's accompanying baseline code will be made publicly available to facilitate open research. The consent of the subjects has been taken following the due process. To the best of our knowledge, this would be one of the first publicly available user engagement datasets being recorded in such diverse conditions. 

\begin{figure*}[t]
\centering
\includegraphics[keepaspectratio=true,width=.95\textwidth]{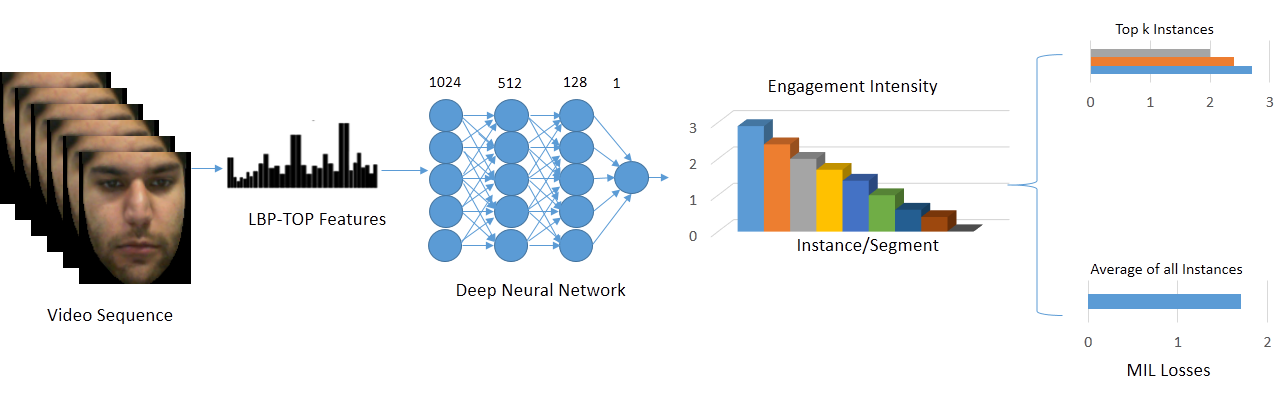}
\vspace*{-3mm}
\caption{\small The proposed deep multi-instance network pipeline. First, LBP-TOP features are extracted from facial video segments. Second linear regression using DNN is employed and instance level responses are ranked. Finally, max and mean pooling are used to predict the subject engagement intensity for a given video.  }
\label{fig:block_diag}
\end{figure*}
\subsection{Data Annotation}
\label{subsec:annotation}
After video recording, the next step is to label the subject\textquotesingle s video for perceived engagement. A team of 5 annotators viewed and rated the videos for the engagement intensity. The labelers were instructed to label the videos on the basis of their engagement intensity (from facial expressions) ranging from 0 to 3. Audio for the videos was turned off during labeling. The engagement categories are inspired by the work of Whitehill \emph{et al.} \cite{whitehill2014faces}. The engagement intensity mapping is as follows: \emph{Engagement intensity 0} means that the subject is completely disengaged, e.g. the subject seems uninterested and looks away from the screen frequently. \emph{Engagement intensity 1} applies to being barely engaged e.g. the subject barely opens his/her eyes, moves restlessly in the chair. \emph{Engagement intensity 2} applies that the subjects seem engaged in the content, e.g. the subject seems to like the content and is interacting with the video. \emph{Engagement intensity 3} means that the subject seems to be highly engaged, e.g. the subject was glued to the screen and was focused.
The distribution of videos in the dataset is as follows 9 videos belong to level 0, 53 for level 1, 82 for level 2 and 50 for level 3. Dataset is split into train and test dataset maintaining subject independence. Train dataset has 147 and test dataset has 48 videos.

For judging the annotator\textquotesingle s reliability, we computed weighted Cohen\textquotesingle s $K$ with quadratic weights as the performance metric (similar to \cite{whitehill2014faces}\cite{gupta2016daisee}) as the labels are not categories but intensities of engagement \{0, 1, 2, 3\}. Any annotator whose agreement coefficient is less than 0.4 is marked less reliable and the label is ignored. After denoising, the labels are averaged and rounded off to the nearest integer to give a ground truth engagement rating to the video. Example images for engagement labels are shown in Fig. \ref{fig:dataset}. The dataset along with the annotations would be made publicly available for the research community.
\section{Proposed Framework}
\label{sec:engagement}
Our framework performs subject's engagement detection and localization. The steps are as follows: The face and facial landmark positions \cite{baltruvsaitis2016openface} are detected in each frame. The video is then sub-sampled into small segments (instances) as discussed in detail in Section \ref{subsection:pre-processing}. Features are extracted for a window of frames. The window then slides, next set of frames are extracted to compute features. This stage is discussed in Section \ref{subsection:lbptop} and \ref{subsection:EandH}. The engagement prediction and localization are computed with the deep MIL network using mean and top-k pooling for regression. The details are discussed in Section \ref{subsection:deepmil_network}. Another model to process the sequence of segments is constructed using Long Short Term Memory(LSTM) based network using mean pooling. The details are discussed in \ref{subsection:deepmil_network}
 
\subsection{Pre-Processing}
\label{subsection:pre-processing}

Let $v= [f_1,f_2,f_3,\cdots,f_n]$ be a video clip of a subject in response to a stimuli, where $f$ denotes a video frame and $n$ are the total number of frames. The OpenFace library \cite{baltruvsaitis2016openface} is used to track the facial landmarks first inorder to register the facial region, which is then cropped to a final resolution of 112 $\times$ 112. The videos in the database are captured at 30 fps, the change in facial expression in consecutive frames can be minimal. Inorder to speed up the experiments, each video is therefore sampled at a frame rate of 6 fps such that $v_s= [f_1,f_7,\cdots,f_\frac{n}{6}]$  represents the sampled video.

\subsection{LBP-TOP Feature Extraction}
\label{subsection:lbptop}

We are interested in computing spatio-temporal features to capture the changes in the facial region. Local Binary Patterns from Three Orthogonal Planes (LBP-TOP) \cite{zhao2007dynamic} is a standard spatio-temporal texture descriptor and is used in our work. The LBP-TOP descriptor concatenates the local binary patterns computed from three orthogonal planes: XY, XT and YT of a video sequence. In the experimental evaluation, LBP-TOP features were extracted in two ways:
\begin{enumerate}
\item \textbf{Full video:} The complete video $v_s$ is considered and LBP-TOP features are extracted from it.

\item \textbf{Shorter segment:} A sliding window $v_f= [f_1,f_2,\cdots,f_k]$ of $k$ frames is taken at a time and LBP-TOP features are extracted. A smaller value of $k$ would fail to efficiently approximate the entire expression while a larger value of $k$ would yield results with poor frame-wise specificity. We therefore deem $k$=20 (approx. 3 seconds in the sampled video) as the appropriate length of each window for expression recognition. 
After extracting features for the first $k$ frames, the sliding window is moved by $l$ frames. The value of $l$ is empirically set to 10 (overlap of 50\%) as it was found to provide adequate accuracy. 
\end{enumerate}
\subsection{Eye Gaze and Head Pose Feature Extraction}
\label{subsection:EandH}
Engagement detection can be inferred from the behavior of a student during learning activities. Engagement levels vary across the learning process. Various changes in body posture and facial moments can be automatically tracked using open source utility softwares. In this paper, social signal in the form of eye gaze movement and head movement features are extracted using OpenFace.  
\begin{enumerate}
\item \textbf{Head Pose}: movement of Head an indicator of engagement intensity. OpenFace returns 6 features related to head movement in the form of x, y and z coordinates as well as roll, yaw and pitch movement of Head.
\item \textbf{Eye Gaze}: The gaze of left and right eye is extracted using Open Face for every frame in the videos.
\end{enumerate}
Intuition to use the head pose and Eye Gaze feature comes from the fact that eye movement gives cue about attention paid by the students in class room environment also. Similarly, Head pose movement also gives cue about the interest of a student while listening to a lecture\cite{d2010multimodal}.

\subsection{Deep Multi-Instance Network for Engagement Localization and Prediction}
\label{subsection:deepmil_network}
In the MIL paradigm the training data is in the form of bags \cite{sikka2013weakly}, $B = \{X_i , y_i \}_{i=1}^{N} $, where $X_i = \{x_{ij}\}_{j=1}^M $, $y_i \in Y$, $M$, the number of instances in $X_i$ and $Y \in \{ 0,1,2,3\}$ are the labels for the bags. $M$ is same across all the videos. Such a problem is common in computer vision where it is easier to label the bag compared to individually labeling the instances. The goal of our work is to efficiently predict engagement intensity for the whole video and localize the most engaging/disengaging segments within a given stimuli video.

We design a deep multi-instance network for the task of user engagement level prediction and localization. Fig. \ref{fig:block_diag} shows the proposed network architecture with multiple dense regression layers, one ranking layer and one multi-instance loss layer. Pooling functions play an important role in MIL deep networks to fuse instance-level outputs. Two pooling schemes are employed for combining multiple instances, \emph{i)} the top k-pooling based multi-instance learning taking only the largest ten elements $(k=10)$ from the ranking layer and \emph{ii)} mean-pooling based multi-instance learning taking mean of all the elements. The details of these schemes are given later in the section.

\subsubsection{Top-k Pooling based Multi-instance Learning}
\label{subsubsec:max_pooling}

A general MIL framework for classification problems defines two kinds of bags, positive and negative. In MIL, if atleast one instance is positive then the bag is assigned a positive label \cite{ dietterich1997solving}. This assumption holds good for binary classification problems like malignancy detection \cite{ zhu2017deep}, pain detection \cite{sikka2013weakly} and image classification \cite{wu2015deep} etc. 

In our dataset, since the ground truth labels are given for the entire video spanning 5 minutes, we take the first $k$ largest intensities of instances from the ranking layer instead of adopting the general multi-instance learning assumption of considering the largest contributing instance. The hyper-parameter $k$ has been chosen from the following set of values $k \in \{ 1, 5, 10, 20, 30\}$ and is empirically set as 10. Furthermore, let $R_i^{'}$ be the rankings of $i^{th}$ bag in descending order such that $R_{i}^{'}= \{r_{ij}^{'}\}_{j=1}^M$, where $r_{ij}^{'}$ is the $j_{th}$ instance in the $i_{th}$ bag. The average of highest $k$ values of the ranking layer \emph{i.e.} $\frac{1}{k}\sum\limits_{j=1}^k r_{ij}^{'} $ is taken as the predicted intensity score for $i^{th}$ bag.

\subsubsection{Mean Pooling based Multi-instance Learning}
\label{subsubsec:mean_pooling}

The engagement intensity fluctuates among various instances over the span of the video. We adopted another scheme of pooling in which the final predicted intensity for $i^{th}$ bag is taken to be the mean of all the intensities from the ranking layer \emph{i.e.}
$\frac{1}{M}\sum\limits_{j=1}^M r_{ij} $.
\subsubsection{Sequence Network}
\label{subsubsec:LSTM}
LSTM based framework is successful in sequence based prediction task. It motivates us to design a network which leverage insights from the sequence of segments. The intuition behind using the LSTM based network is that engagement level varies across the segments. The annotators gives attention to different segments of the recorded video and gives label on the basis of perceived engagement level. It implies different segments contribute differently to the overall annotation of the video.\\
The figure 3 presents the proposed architecture of the sequence based architecture. In this network the first layer is LSTM layer which has 32 hidden units. The purpose of this layer is to compute an activation value for the sequence of segments of a video. This maps the multi segment representation of a video to a the feature vector which represents the activation for different segments of the video.
It is followed by flatten layer which serves the purpose to make a single feature vector of activations of all the segments of the video.
This feature vector is passed to another module which consists of three dense layers with sigmoid activations followed by average pooling. This module is responsible to learn the function which maps the segment level activations to the engagement level of a video.

\begin{figure*}[t]
\centering
\label{figure:LSTMW}
\vspace*{-35mm}
\includegraphics[keepaspectratio=true,width=1.2\textwidth]{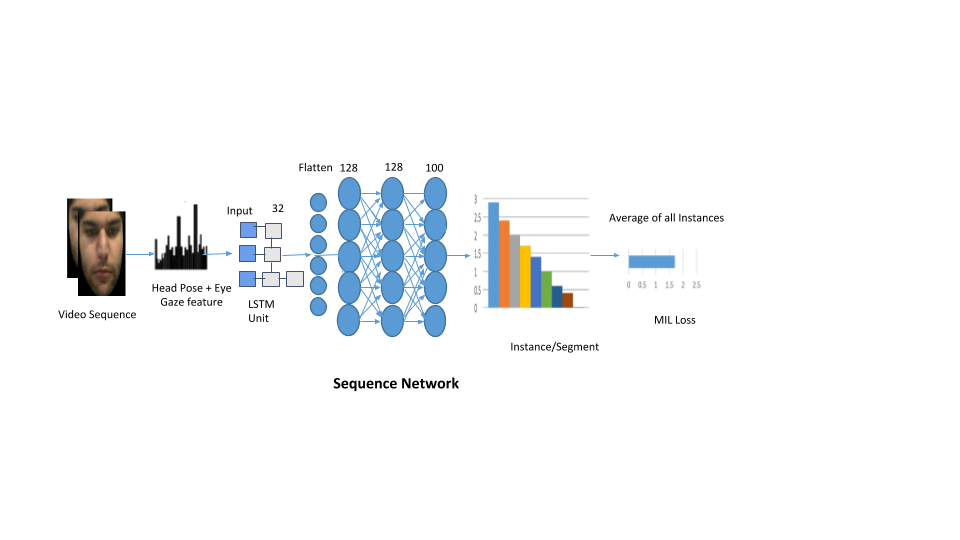}
\vspace*{-35mm}
\caption{\small The proposed deep multi-instance network pipeline. First, Eye-Gaze and Head Pose features are extracted from facial video segments. Second, the sequence of segments is passed through a LSTM layer followed by 3 dense layers, finally average pooling layer is applied. }
\label{fig:block_diag_lstm}
\end{figure*}
\section{Experiments}
\label{sec:experiments}
The details about the dataset which was used for experimentation can be found in Section \ref{sec:data}. We followed subject independent cross-validation for dataset split. The training and test sets are also subject independent. In this section, we discuss the two approaches followed to predict the engagement level of individuals and result comparisons are shown in Section \ref{sec:results}.

\subsection{Multiple Instance Learning Approach}
\label{subsec:traditional_ml}

 We have explored two approaches of extracting LBP-TOP features as explained in Section  \ref{subsection:lbptop}. In first approach, features for the entire video are extracted as a single vector $\{x_i\}_{i=1}^N$. The training data over $N$ samples is represented as $D = \{x_i , y_i \}_{i=1}^{N} $, where $x_i \in X$ and $y_i \in Y$. 

In second approach a temporal scanning window is taken and features are extracted for each window segment of the video. The training data is in the form of bags, $B = \{X_i , y_i \}_{i=1}^{N} $, where $X_i = \{x_{ij}\}_{j=1}^M $, $y_i \in Y$ and $M$   (number of instances per video) is equal to 100 and $N$ is the total number of videos in the training dataset.

The segments obtained for each sequence are labeled using the following two approaches:

\subsubsection{Noisy/ Weak Labeling}
\label{subsubsec:segment_label}

All the segments in the training data are assigned the same label as that of their corresponding video sequence. This strategy is similar to that employed in previous works \cite{ashraf2009painful} \cite{lucey2008improving}.
\subsubsection{Data Augmentation}
In our dataset, the proportion of classes labeled as 0 is very less. Only 9 videos are annotated as 0 in the dataset. So in the training dataset for 0 label data augmentation is required to increase the sample points. In our case we  repeated the 0 label videos of training dataset 20 times and 3 label class 2 times as a result the sample points of all the classes are proportional.
\subsubsection{Relabeling segments using Kmeans clustering}

In this approach, the segments are divided into $k$ clusters using the Kmeans clustering over the features. The hyperparameter $k$ is selected from the following set of values $k \in \{ 5, 10, 15, 20, 25, 30, 35, 40\}$. Segments falling in a cluster are relabeled as follows: 

\emph{i)} All the segments in a cluster are relabeled by the mode of the labels in that cluster. 
\emph{ii)} All the segments in the cluster are relabeled by the mean of the labels in that cluster. 
\subsubsection{Regressors} Several state of the art machine learning models are trained for the regression task.
\emph{i)}\emph{ Support Vector Regression (SVR)}: This classifier works on the relabeled segments and regress the value of the video as the mean value of all the segments belonging to that particular video. A Gaussian kernel is trained by performing a grid search on the selection of parameters C and sigma. Mean squared error is obtained for the best combination of C and Sigma for different mode of segment relabeling techniques as mentioned earlier. For mode analysis C and Sigma is (1,1) whereas for mean analysis  C and Sigma is (1,4).  
Other linear regressors are also compared. \emph{SGDREgressor:} which fits linear regression with specified loss function as mean squared error and penalty as 0.0001 and
\emph{ BayesianRidge Regressor} is used to regress the engagement level of a video from the regressed values of its segments with different parameter settings.   \\
As shown in Table \ref{table:Results_traditional}, 
the different regressors are used to regress the engagement level using segments of a video. It is using linear regression, DNN with different relabeling techniques. It can be observed the traditional Multi Instance Learning (MIL) methods could not learn the complex function required for regression in this setting. 

\subsection{Deep Multi Instance Learning}
\label{subsec: deep_mil_experiments}
\emph{DNN: }
The architecture for Deep MIL approach is shown in Fig. \ref{fig:block_diag}. Here a video sequence is first divided into 100 segments and LBP-TOP features are extracted from each segment. These features are then passed to a neural network with four dense layers, obtaining a single feature for each segment. To obtain a single label for the entire video, segment-wise features are then passed through a Deep MIL loss layer using two pooling schemes, namely Top-k and Mean as mentioned in Section \ref{subsection:deepmil_network}.
 Top-k pooling, intensities of the top $k$ contributing instances are evaluated and their mean is finally assigned as the predicted intensity value for the video. In mean pooling, the average of intensities of all the instances of the video is assigned as the final intensity for the video. 

We also evaluated Rectified Linear Unit (ReLU): $f(x) = max (0,x)$ ,activation function the most frequently used activation function as it allows gradient flow for all positive values.  

\emph{Proposed Method: }The architecture for LSTM based Deep MIL approach is shown in Fig. \ref{fig:block_diag_lstm}.
Here, the video is divided into segments. each segment is represented as standard deviation of x, y and z coordinates of frames present in a segment. These points are returned by Open Face.
The eye gaze movement is also represented as standard deviation of the points returned for gaze of left and right eye in a particular segment the video. As a result, both the eye and head pose features are concatenated resulting in 9 dimensional feature vector. Each video is represented using collection of segments where each segment is represented as a fused feature having information of head pose and eye gaze. These features are passed through the LSTM layer which returns activation for each segment of the video, passed to the flatten layer and then flattened feature vector is passed to the network of three dense layers followed by average pooling which gives the regressed value of engagement level of a video.

\section{Results and Discussions}
\label{sec:results}

In this paper, we have presented the efficacy of DMIL in handling multiple segments in a video compared to the baseline ML algorithms. Our results show a significant performance improvement using Deep MIL approach to affirm our proposition. Table. \ref{table:Results_traditional} shows the mean squared error in case of LBPTOP features, for baseline ML approaches and gives a clear view of the different relabeling techniques discussed in Section \ref{subsec:traditional_ml}. It is observed that different relabeling techniques have different performance.
For other linear regressors, the error is substantially more than SVR. It implies the linear function is not a suitable choice for this problem

Table \ref{table:Results_DeepMIL } shows the performance of our network using mean squared error, Pearson\textquotesingle s correlation coefficient (PCC) as measures for the strength of correlation between annotated and regressed engagement level. Our results show the relevance of measuring classwise mean square error. Although, the mean square error reduced overall but the correlation coefficient is very less it implies alone mse is not sufficient to measure the accuracy of the network to regress the value of engagement level.\\
Table \ref{table:Results_DeepMIL_1 }, shows the performance of our network using mean squared error for each engagement level for LBPTOP features. It can be observed network only  with dense layers (DNN) is not able to capture the class wise accuracy and it is not performing well for level 0 and level 1. It is because it is has regressed values very close to each other for all the test instances. In case of LSTM based proposed network, LBPTOP features are not performing well.\\
Table \ref{table:Results_traditional headpose}, shows the performance of fused features of Head Pose and Eye Gaze with SVR and different relabeling techniques of the segment. It can be observed that mse error level 0 is high. In all the cases, it can be said traditional methods with fused features is not appropriate choice for engagement level measurement.\\
Table 6, shows the performance of Fused features for both DNN and Sequence network using mean squared error for each engagement level. It can be observed that mean pooling with Relu activation applied on dense layers following the activations returned by LSTM layer performs better than any other network. \emph{This network is able to minimize the class wise MSE and overall MSE is also less in comparison. The predicted labels are more correlated with the annotated data as PCC has increased to 0.37.}\\
\textbf{User Engagement Localization: }
Instance level engagement intensities $R_i = \{r_{ij}\}_{j=1}^M$ corresponding to $X_i = \{x_{ij}\}_{j=1}^M$ (see Section \ref{subsection:deepmil_network}) are extracted from the penultimate layer of proposed network. In Fig. 5, we show 4 engagement levels to highlight the ability of our algorithm to localize engagement.
Every point $i$ along the X-axis denote the  penultimate layer output of the Sequence network based network, for $i^{th}$ segments of the video belonging to a particular subject and engagement level.\\
In Fig. 4, The average of all the segements of the videos of the same level is plotted and
\emph{It can be observed that the Sequence network (proposed network) using fused features of Head Pose and Eye Gaze is able to differentiate between the (level 0) and (level 1, level 2, level 3) very nicely which implies engaged and non engaged videos can be differentiated with the proposed network. It is not able to correctly distinguish at fine grain level i.e between levels 2 and 3.} Reason for the same could be noise in the features space as well as the feature space has overlapping feature distribution for different engagement levels. Our experiments are likely to show promising results in applications where the goal is to foster student engagement e.g. in MOOCs and ITS.
\begin{figure*}
\centering
\label{figure:eng}
\vspace*{-5mm}
 \includegraphics[keepaspectratio=true,width=0.5\textwidth]{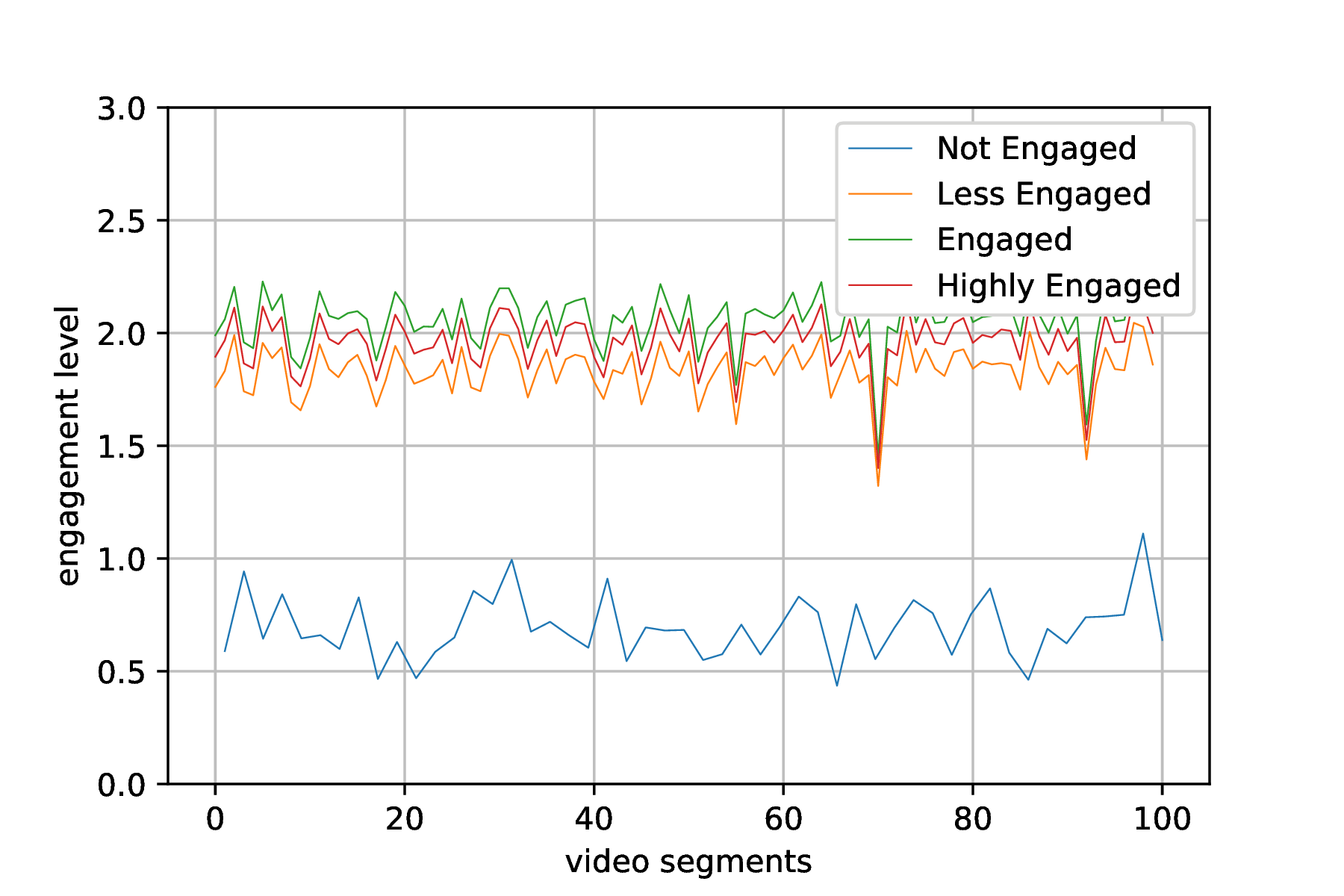}
\vspace*{-3mm}
\caption{\small This figure represents the predicted engagement for the segments average over the videos of the same label.For Level 0 most of the predicted values of engagement are between 0 and 1. For level 1 the average is going high so the mse error for this class is high whereas for level 2 it is near the ground truth. Level 3 is not predicted with high accuracy so the average is below ground truth.  }
\label{fig:avg_eng}
\end{figure*}
\label{sec:experiments}

\begin{table}
\caption{Baseline performance comparison of different ML approaches for LBPTOP. Here Relabeling refers to the method of segment-wise labeling.} \vspace{-2mm}
\label{table:Results_traditional}
\centering
\begin{center}
\vspace{-2mm}
\begin{tabular}{|c||c|c|}
\hline 
\textbf{Method} & \textbf{Relabeling} & \textbf{Error} 
\\
\hline \hline
SVR & Noisy labeling & 0.15  \\ 
\hline 

SVR & mode & 0.15 \\
\hline

SVR & mean & 0.15  \\
\hline

SGDRegressor & mean & 1.20  \\
\hline

BayesianRidge& mean & 1.20 \\
\hline



DNN& Mean  & 0.15\\
\hline
\textbf{LSTM}& \textbf{Mean} & \textbf{0.10} \\
\hline
\end{tabular}
\end{center}  \vspace{-9mm}
\end{table}

\begin{table}
\caption{Results obtained from LBPTOP for different DMIL methods. } \vspace{-2mm}
\label{table:Results_DeepMIL }
\centering
\begin{center}
\begin{tabular}{|c||c|c|c|}
\hline 
\textbf{Method} & \textbf{Pooling } & \textbf{Error} &  \textbf{PCC} \textbf{(R)}\\ 

\hline \hline
DNN& Max k   & 0.16& 0.003 \\ 
\hline 
DNN& Mean  & 0.15 & 0.0264 \\ 
\hline
\textbf{LSTM}& \textbf{Mean} & \textbf{0.10}& \textbf{0.001} \\
\hline

\end{tabular}
\end{center}

\end{table}

\begin{table}
\caption{Classwise MSE results obtained from DMIL using LBPTOP features. } 
\label{table:Results_DeepMIL_1 }
\centering
\begin{center}
\begin{tabular}{|c||c|c|c|c|c|}
\hline 
\textbf{Method} & \textbf{Pooling } & \textbf{Level 0} &  \textbf{Level 1}& \textbf{Level 2} & \textbf{Level 3}\\

\hline \hline
DNN&Max k  & 0.53& 0.20 & 0.053&0.14\\ 
\hline 
DNN&Mean & 0.33 & 0.14 & 0.03&0.21\\ 
\hline
\textbf{LSTM}&\textbf{Mean}  & \textbf{0.36}& \textbf{0.09} & \textbf{0.004}&\textbf{0.15}\\
\hline

\end{tabular} 
\end{center}
\end{table}

\begin{table}
\caption{Baseline Performance comparison of different classifiers for Fused feature of Head Pose and Eye Gaze} 
\label{table:Results_traditional headpose}
\centering
\begin{center}
\begin{tabular}{|c||c|c|c|c|c|c|}
\hline 
\textbf{Method} & \textbf{avg. MSE}& \textbf{ level 0 } & \textbf{level 1} &  \textbf{level 2}& \textbf{level 3 }  \\
\hline
\textbf{SVR+Noisy}  &\textbf{ 0.09} &\textbf{0.38}  &\textbf{0.11} &\textbf{0.003}&\textbf{0.12} \\
\hline

SVR+Mode &0.15  &0.79  &0.31 &0.05&0.01 \\
\hline
SVR+Mean & 0.15& 0.77& 0.31&0.05&0.01 \\
\hline

\end{tabular}
 \end{center}  
 \vspace{-3mm}
\end{table}
\begin{table} 
\caption{Performance comparison of DMIL approaches for the fused features of Head Pose and Eye Gaze} 
\label{table:LSTM model}
\centering
\begin{center}
\begin{tabular}{|c|c|c|c|c|c|c|}
\hline 
\textbf{Method} & \textbf{avg. MSE}& \textbf{ Level 0 } & \textbf{Level 1} &  \textbf{Level 2}& \textbf{ Level 3 }  \\

\hline
DNN &0.11  &0.20  &0.15 &0.02&0.18 \\
\hline
\textbf{LSTM} & \textbf{0.10}& \textbf{0.20}& \textbf{0.16}&\textbf{0.03}&\textbf{0.08} \\
\hline

\end{tabular}
\end{center} 
\end{table}
\begin{table}
\caption{Results of Mean pooling for Fused Features and correlation coefficient of predicted Engagement levels} 
\label{table:LSTM model}
\centering
\begin{center}
\begin{tabular}{|c|c|c|c|}
\hline 
\textbf{Method} & \textbf{Pooling }&\textbf{avg mse}& \textbf{ PCC}  \\

\hline
DNN &mean &0.11  & 0.07  \\
\hline
\textbf{LSTM} &\textbf{ mean} &\textbf{0.10}& \textbf{0.25} \\
\hline

\end{tabular}
\end{center} 
\vspace{-6mm}
\end{table}

\begin{figure*}
\begin{multicols}{2}
   \vspace*{-6mm} \includegraphics[keepaspectratio=true,width=0.4\textwidth]{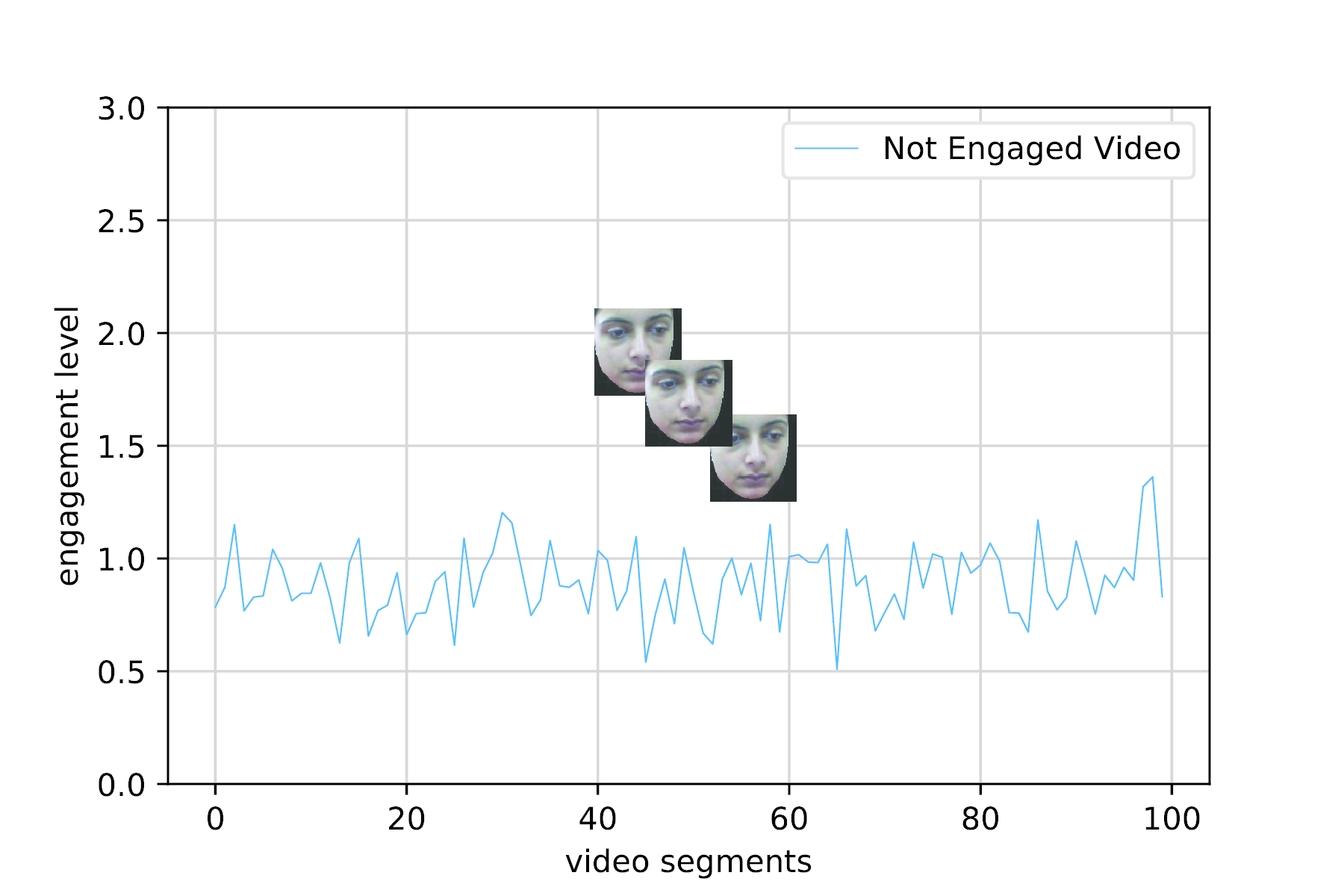}\par 
    \includegraphics[keepaspectratio=true,width=0.4\textwidth]{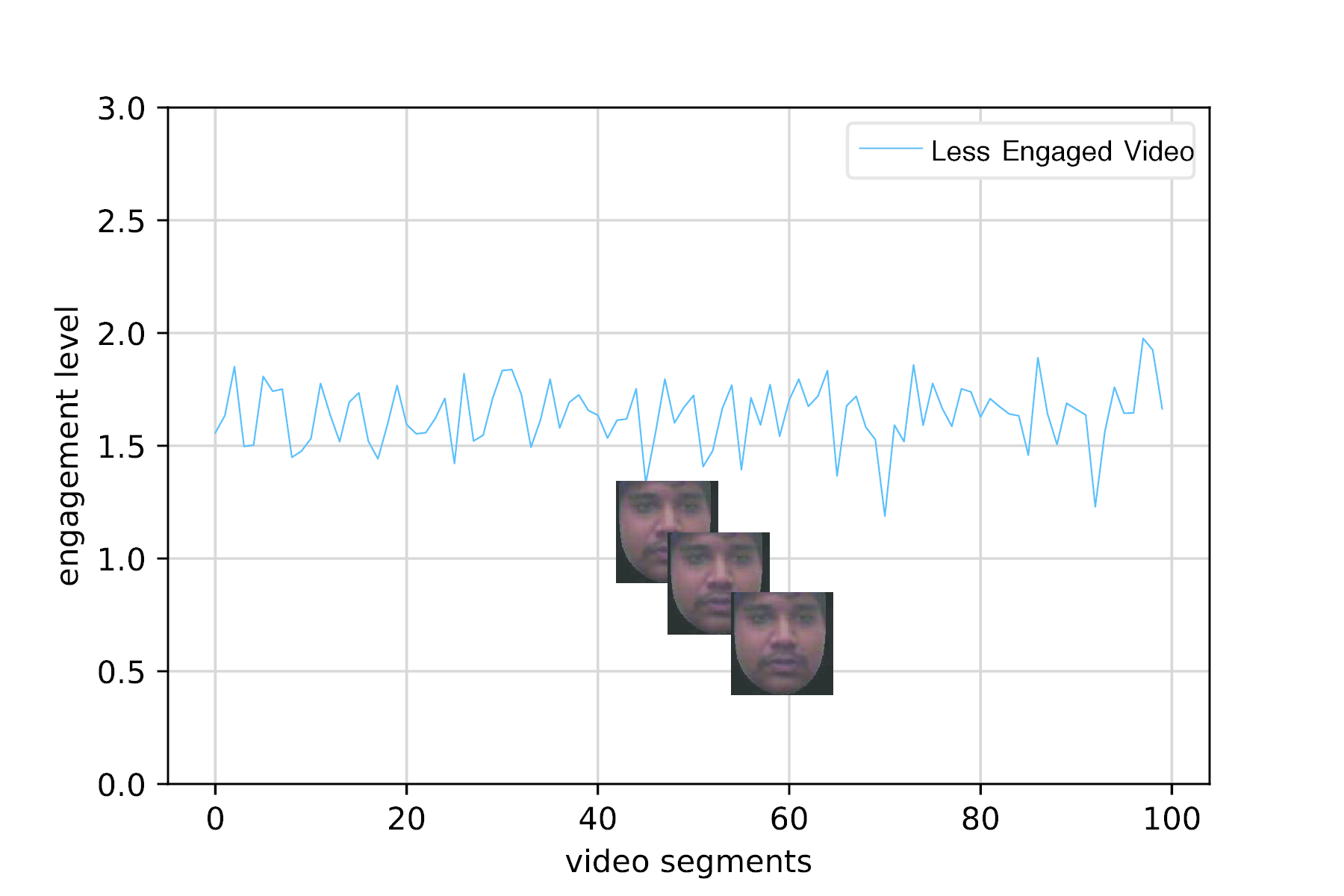}\par 
    \end{multicols}
    \vspace*{-12mm}
\begin{multicols}{2}
    \includegraphics[keepaspectratio=true,width=0.4\textwidth]{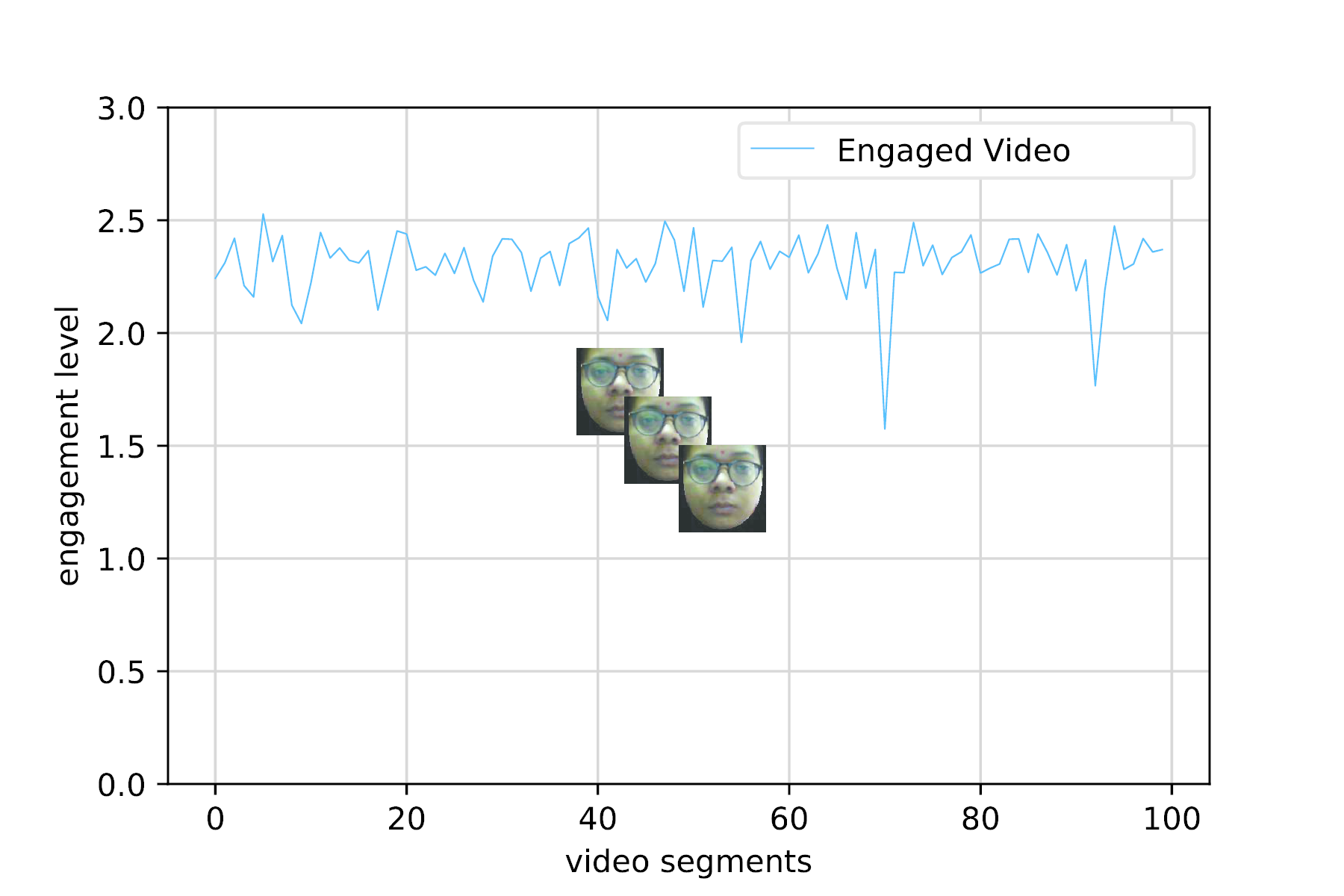}\par
    \includegraphics[keepaspectratio=true,width=0.4\textwidth]{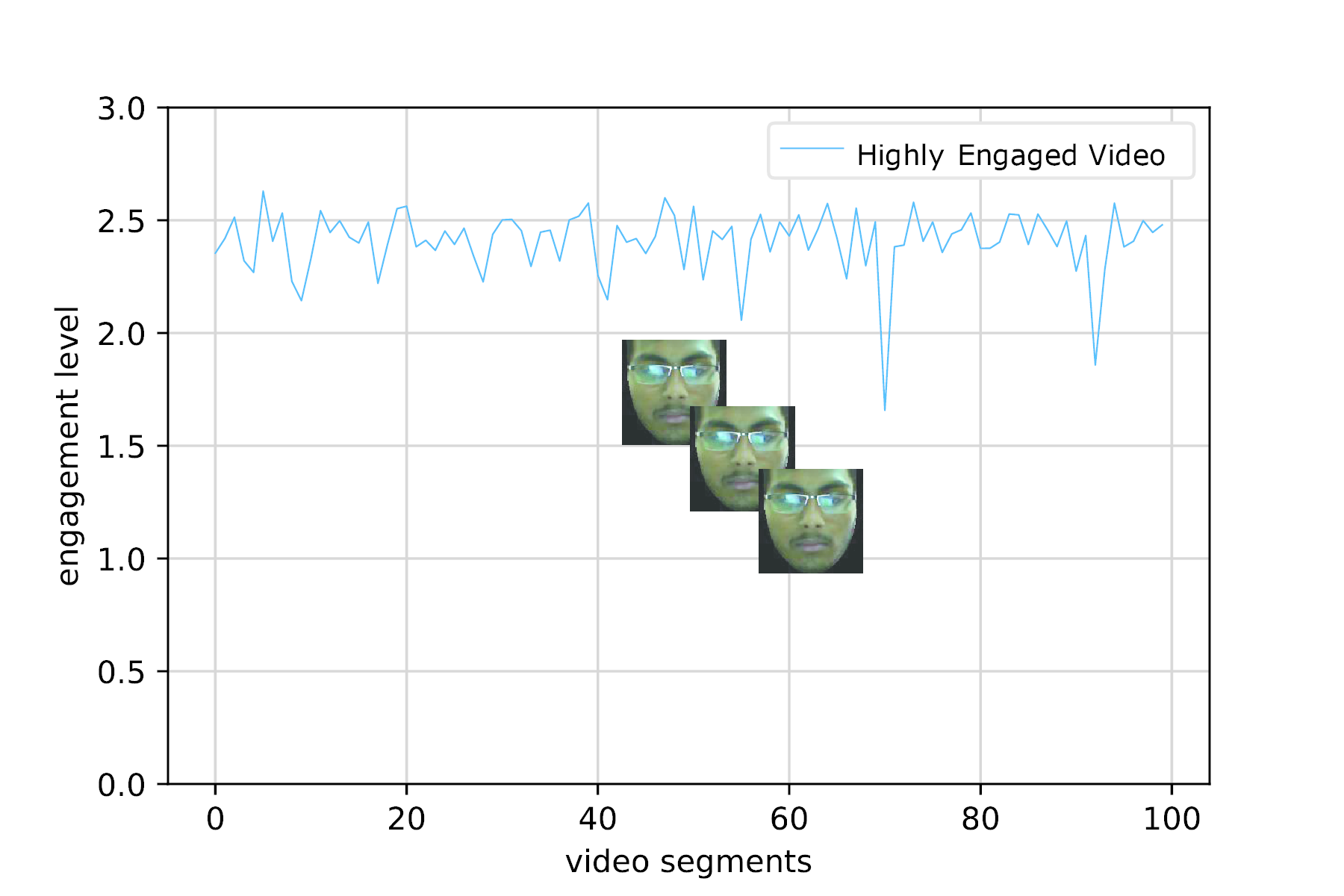}\par
    
\end{multicols}
\caption{\small Examples showing predicted engagement for the segments of the videos of the different level. For Level 0 most of the predicted values of engagement are between 0 and 1. For level 1 and 2 it is near the ground truth. Level 3 is not predicted with high accuracy and same is depicted in the diagram. }
\vspace*{-5mm}
\end{figure*}
\section{Future Work}
In this work, we have demonstrated the effectiveness of Sequence Network to detect engagement level based on the engagement of the segments. In future, we are planning to consider the body movement feature vector in the form of optical flow and movement of different parts of body.  Further, to the best of our knowledge this is the first attempt to localize student engagement over the course of a video. In this work the students from the institute of different department are employed whereas in future the non institute people of older age can also be recruited. This can give more freedom for generalization of the results.
\section{Conclusion}
\label{sec:conclusion}
In automatic engagement, many of the current techniques used- like self-reports, teacher introspective evaluation are cumbersome and lack the ability to localize student engagement. In this paper, we have devised a multi-instance learning-based method to automatically predict the engagement level of a subject from the motion analysis in the form of movement of head pose and eye gaze movement. In prior literature, other modalities such as student response, audience movie rating, psychological and neurological sensor readings etc (see Section \ref{sec:intro}) have been used for engagement prediction, however, our deep MIL based approach along with sequence network performs better then other networks. In this work LBPTOP has not performed well for this in the wild dataset whereas motion based feature vector are able to detect and localize the engagement levels. Sequence network is powerful to detect high level classification of engaged and non engaged videos but is not so effective in fine grain classification of level 2 and level 3 videos. The proposed method can be extensively used in designing study material for MOOCs and ITS to substantially decrease the dropout rates.

\section{ACKNOWLEDGEMENTS}
We would like to thanks NVIDIA for donating Titan X for our research work.

\pagebreak
\newpage
\bibliographystyle{ACM-Reference-Format}
\bibliography{ref.bib}

\end{document}